\documentclass{article}

\usepackage[preprint]{neurips_2024}

\usepackage[utf8]{inputenc} 
\usepackage[T1]{fontenc}    
\usepackage{hyperref}       
\usepackage{url}            
\usepackage{booktabs}       
\usepackage{amsfonts}       
\usepackage{nicefrac}       
\usepackage{microtype}      

\usepackage{comment}
\usepackage{amsmath}
\usepackage[ruled,vlined]{algorithm2e}
\usepackage{algpseudocode}
\usepackage{amsthm}
\usepackage{amssymb}
\usepackage{enumitem}
\usepackage[table]{xcolor}
\usepackage[export]{adjustbox}
\usepackage{wrapfig,lipsum,booktabs}
\usepackage{makecell}
\usepackage{caption}
\usepackage{mathrsfs}

\usepackage{multirow}
\usepackage{arydshln}
\usepackage{bm}
\newcommand{\cmark}{\ding{51}}

\usepackage{pifont}
\usepackage{enumitem}
\usepackage{amssymb}
\usepackage{pdfpages}

\def\etal{\textit{et~al. }}

\definecolor{mygray}{gray}{.9}
\definecolor{mygray1}{gray}{.97}
\definecolor{mygreen}{RGB}{93,173,85}
\newcommand{\pub}[1]{{\color{gray}{\tiny{[{#1}]}}}}

\newcommand{\model}{\textsc{ReliOcc}\xspace}

\makeatletter
\newcommand{\thickhline}{
    \noalign {\ifnum 0=`}\fi \hrule height 1pt
    \futurelet \reserved@a \@xhline
}



\title{\model: Towards Reliable Semantic Occupancy Prediction via Uncertainty Learning}

\author{Song Wang$^{1}$, Zhongdao Wang$^{2}$, Jiawei Yu$^{1}$,  Wentong Li$^{1}$,  \\  \textbf{Bailan Feng$^{2}$,~~Junbo Chen$^{3}$,~~ Jianke Zhu$^{1}$\thanks{Corresponding author.}} \\
  $^{1}$Zhejiang University \quad
  $^{2}$Huawei Noah's Ark Lab  \quad
  $^{3}$Udeer.ai
  }

\begin{document}

\maketitle

\begin{abstract}
Vision-centric semantic occupancy prediction plays a crucial role in autonomous driving, which requires accurate and reliable predictions from low-cost sensors. Although having notably narrowed the accuracy gap with LiDAR, there is still few research effort to explore the reliability in predicting semantic occupancy from camera. In this paper, we conduct a comprehensive evaluation of existing semantic occupancy prediction models from a reliability perspective for the first time. Despite the gradual alignment of camera-based models with LiDAR in term of accuracy, a significant reliability gap persists. To addresses this concern, we propose \model, a method designed to enhance the reliability of camera-based occupancy networks. \model provides a plug-and-play scheme for existing models, which integrates hybrid uncertainty from individual voxels with sampling-based noise and relative voxels through mix-up learning. Besides, an uncertainty-aware calibration strategy is devised to further enhance model reliability in offline mode. Extensive experiments under various settings demonstrate that \model significantly enhances model reliability while maintaining the accuracy of both geometric and semantic predictions. Importantly, our proposed approach exhibits robustness to sensor failures and out of domain noises during inference.
\end{abstract}

\section{Introduction}
The goal of semantic occupancy prediction is to obtain a comprehensive voxel-based representation of the 3D scene from either LiDAR point clouds~\cite{roldao2020lmscnet, xia2023scpnet} or camera images~\cite{cao2022monoscene, huang2023tri, li2023voxformer}, which is crucial for the perception systems in autonomous driving and robotics. 
Initially, LiDAR-based models~\cite{roldao2020lmscnet,cheng2021s3cnet,yan2021sparse,xia2023scpnet} dominated the field due to their ability to provide accurate geometric cues.
Researchers nowadays prefer to learn 3D occupancy information from images owing to the low cost and widespread availability of camera sensors.
Recent progress~\cite{cao2022monoscene, huang2023tri, li2023voxformer, yao2023ndc, mei2023camera} has significantly narrowed the gap in accuracy between camera and LiDAR-based approaches. 
However, their performance in terms of reliability remains under-explored, which becomes paramount in safety-critical scenarios.

Traditionally, the occupancy labeling relies on accumulated LiDAR point cloud and its corresponding 3D semantic labels~\cite{behley2019semantickitti,tian2024occ3d,wang2023openoccupancy,wei2023surroundocc}. 
With the development of vision-centric approaches~\cite{cao2022monoscene,huang2023tri,yao2023ndc,li2023voxformer,mei2023camera} using images, questions arise regarding the reliability of predictions solely derived from cameras without accurate depth information. 
Since the overall accuracy of occupancy networks is relatively low, exploring the reliability and uncertainty of their predictions can provide valuable reference information for downstream tasks, such as decision-making and planning~\cite{zheng2023occworld, hu2023planning, albrecht2021interpretable}.

With the above considerations, we conduct a thorough evaluation of existing semantic occupancy prediction models based on a reliability standpoint. 
To achieve this, we introduce the misclassification detection and calibration metrics from both geometric and semantic dimensions for evaluating model that utilize camera or LiDAR data. 
Our findings reveal that, camera-based models often lag behind their LiDAR-based counterparts in terms of reliability despite improvements in accuracy.

To mitigate this disparity, \model is proposed to improve the reliability in occupancy networks via a new hybrid uncertainty learning scheme. 
Our approach optimizes uncertainty by taking into account of perturbations in individual voxels (\textit{absolute uncertainty}) and the relative relationships in mix-up voxels (\textit{relative uncertainty}) during model training.
By integrating multiple sources of information for uncertainty learning, our method enhances the reliability of camera-based models without sacrificing inference speed or accuracy. 
Moreover,  we provide an uncertainty-aware calibration strategy to utilize the learned uncertainty in offline mode, further enhancing model reliability.
Through extensive experiments across diverse configurations including online and offline modes, our method achieves competitive performance compared against the state-of-the-art models.

Our main contributions can be summarized as follows: 
\begin{itemize}[leftmargin=*]
    \item A comprehensive evaluation is conducted on existing semantic occupancy prediction models from a reliability perspective, which provides a series of misclassification detection and calibration metrics across both geometric and semantic dimensions.
    \item To adapt the existing methods for occupancy networks, we propose the \model to enhance the reliability of camera models. A novel hybrid uncertainty learning approach is presented to combine the variance from individual and mix-up voxels and significantly narrow the reliability gap between camera and LiDAR-based methods. 
    \item Extensive experiments on online uncertainty learning and offline model calibration under various settings, demonstrate the effectiveness of our approach on general condition and robustness in adverse condition like sensor failures and noisy observations.
\end{itemize}

\section{Preliminaries}
\label{sec:prelim}
\subsection{Problem Formulation}

\noindent\textbf{Occupancy Prediction.} Given inputs $\bm{x}$ from LiDAR or camera sensors, occupancy networks $V_\theta (\bm{x})$ generate dense features $\mathcal{V} \in \mathbb{R}^{ d\times L \times W \times H}$ in a pre-defined volume, where $L$, $W$, and $H$ represent the length, width, and height, respectively. $d$ is the dimension of dense features.
For any voxel $v_i \in \mathbb{R}^{d}$ within this volume, the prediction involves with two components. One is a binary indicator that specifies whether the voxel is occupied or not. The other is the semantic label of the voxel if the voxel is occupied. Generally, such process can be formulated by estimating the probability $p(y_i=y|v_i)$ for $v_i$, where $y \in \{0, 1, ..., S\}$. Here, $0$ denotes that the voxel is unoccupied, and $S$ is the total number of semantic classes.

\noindent\textbf{Misclassification Detection.} 
For a reliable classifier, we expect it to accurately reject those incorrect predictions with low-confidence. Misclassification detection is introduced to measure the gap between the actual trained model and the ideal one, which can be evaluated by \textit{rejection curves}~\cite{fumera2002support, hendrycks2021natural}.
To avoid the tendency of higher precision models, we adopt the same strategy as ~\cite{malinin2019ensemble,de2023reliability} to normalize the area under the curve and deducts a baseline score.

\noindent\textbf{Calibration.} As a long-standing problem in machine learning, the goal of model calibration is to ensure that predicted confidence of a model aligns accurately with the actual likelihood of correctness~\cite{niculescu2005predicting, guo2017calibration}, thereby producing more reliable predictions. Within the framework of multi-class classification, a model is deemed perfectly calibrated if $p(y_i=y|c_i=c, v_i) = c$. Here, the model not only predicts a discrete label $y$ but also generates a confidence score $c \in [0,1]$. This score $c$ should ideally reflect the true probability that the predictions are correct.

\subsection{Evaluation Metrics}
\label{eval-metric}

\noindent\textbf{Canonical Metrics.} 
Following common practice~\cite{song2017semantic, behley2019semantickitti}, we employ the Intersection over Union (IoU) metric to assess the accuracy of geometric occupancy prediction that is typically treated as a binary classification task. 
Additionally, we utilize the mean Intersection over Union (mIoU) across multiple categories to evaluate the quality of semantic predictions. These two metrics are calculated using discrete predictions by applying \texttt{argmax} operation to the logits. Although IoU and mIoU effectively reflect the performance of model in accuracy, they do not assess its reliability.

In this paper, we mainly evaluate the reliability of occupancy prediction on two aspects including misclassification detection and calibration, which are assessed with following metrics.

\noindent\textbf{Prediction Rejection Ratio (PRR).}
The Prediction Rejection Ratio (PRR)~\cite{malinin2019ensemble} is defined through \emph{rejection curves} for misclassification detection. 
To construct a rejection curve, we initially sort predictions based on a specific criterion, such as predicted confidence or oracle confidence (where predictions are labeled 1 if correct and 0 otherwise). Subsequently, a threshold is set and predictions below this threshold are rejected, allowing us to calculate a rejection rate.
As this threshold is incrementally adjusted, we obtain a rejection curve to illustrate how the classification error (depicted on the \textit{y}-axis) decreases in tandem with the rejection rate (represented on the \textit{x}-axis). The PRR metric is then quantitatively defined as follows
\begin{equation}
\texttt{PRR} = \frac{AUC_{\texttt{random}} - AUC_{\texttt{uncertainty}}}{AUC_{\texttt{random}} - AUC_{\texttt{oracle}}},
\end{equation}
where AUC represents the Area Under the Curve. Here, $AUC_{\texttt{random}} = 0.5$ corresponds to the AUC for randomly generated confidences. A perfectly reliable model would achieve a $\texttt{PRR} = 1$. For occupancy networks, we report both $\texttt{PRR}_{\texttt{geo}}$ for geometric predictions and $\texttt{PRR}_{\texttt{sem}}$ for semantic predictions, respectively.

\noindent\textbf{Expected Calibration Error (ECE).}
Expected Calibration Error (ECE)~\cite{naeini2015obtaining, guo2017calibration} assesses the calibration of probabilistic predictions made by machine learning models. It measures the difference between predicted probabilities and observed frequencies across various confidence levels. Intuitively, 
\begin{equation}
\label{eq:ece}
    e_\texttt{ECE} = \mathbb{E}_{\hat{c}_i}[~ |~ p(\hat{y}_i = y_i ~|~ \hat{c}_i = c) - c ~| ~]~.
\end{equation}
A perfectly calibrated model yields $e_\texttt{ECE}=0$.
Eq.~\eqref{eq:ece} is a continuous integration over $c \in [0,1]$. Practically, we approximate this integration by discretizing $c$ into $M$ small bins. Denoting the set of samples falling into the $m$-th bin as $B_m$, the expectation can be calculated as 
\begin{align}
    \texttt{ECE} =  \sum^M_{m=1} \frac{|B_m|}{N} \left|~ \text{acc}(B_m) - \text{conf}(B_m) ~\right|~,
\label{eq:ece3d}
\end{align}
where acc$(\cdot)$ denotes the mean accuracy, and conf$(\cdot)$ is mean confidence of $B_m$. $N$ is the number of samples. We set the number of bins $M= 15$ by default. As with \texttt{PRR}, we report both $\texttt{ECE}_{\texttt{geo}}$ and $\texttt{ECE}_{\texttt{sem}}$ for geometric and semantic predictions, respectively.

\section{Adaptation with Existing Methods for Occupancy Networks}
\label{sec:exist-method}
Reliable predictions are paramount in occupancy networks, especially in critical applications such as autonomous driving and robotics where safety is a strict requirement. 
Despite their importance, methods for enhancing the reliability of occupancy networks are still under-explored in the existing literature. 
To address this gap, we begin by reviewing existing uncertainty learning and calibration methods, which are mostly developed to improve the reliability for traditional tasks. Then, we adapt them for the recent occupancy networks.

We categorize these methods into two paradigms. One is training uncertainty predictor $c_{\sigma|\phi}$ based on the dense features $\mathcal{V}$ concurrently with $V_\theta$ from scratch, which is termed \emph{online uncertainty learning}. Another is training scaling factor $c_{f|\phi}$ on top of a fixed $V_\theta$, which is termed \emph{offline model calibration}.
In the experimental section (see \S\ref{exp:unc-learning} and \S\ref{exp:model-calib}), we provide extensive evaluations on these methods to compare their effectiveness in boosting the reliability of occupancy networks.

\subsection{Online Uncertainty Learning}
\label{sec:3unc-learn}
Uncertainty estimation is a long-standing problem in the context of Bayesian deep learning~\cite{tishby1989consistent,denker1990transforming,gal2016dropout}.  
Prior arts can be classified into ones concerning epistemic (model) uncertainty~\cite{zhou2022survey, jungo2019assessing} and ones concerning aleatoric (data) uncertainty~\cite{hullermeier2021aleatoric, kendall2017uncertainties}.
Although explicit uncertainty estimates are obtainable, we do not directly evaluate these estimates in online mode. 
Instead, since the uncertainty is learned concurrently with the model's predictions from scratch, we use them as a regularization term of helping the model become more reliable. 

For each voxel feature $\mathbf{v_i}$, we compute a logit vector  $\mathbf{z_i} \in \mathbb{R}^{S+1}$ using a linear layer, where $S$ represents the number of semantic classes.

\noindent\textbf{Heteroscedastic Aleatoric Uncertainty (HAU)}
~\cite{kendall2017uncertainties}
is a data-dependent uncertainty learning method. 
We employ the classification form of HAU, which modifies upon a deterministic model by placing a Gaussian over the logit:
$
    \hat{\mathbf{z}}_i | \phi \sim \mathcal{N}(\mathbf{z}_i, (\sigma_i^\phi)^2)
$\footnote{We omit predicting the mean $\mathbf{\mu}^{\phi}(\mathbf{z}_i)$ and use $\mathbf{z}_i$ for simplicity. Empirical results are similar.}. 
The sampled logit vector $\hat{\mathbf{z}}_i$ is then passed through a \textit{softmax} operator and cross entropy loss is computed. Here $\sigma_i^\phi$ is the predicted uncertainty parameterized by $\phi$. 
Optimization of $\phi$ can be done with back-propagation using the re-parameterization trick~\cite{kingma2013auto}: $\hat{\mathbf{z}}_i = \mathbf{z}_i + \sigma_i^\phi \epsilon, \epsilon \in \mathcal{N}(\mathbf{0}, \mathbf{I})$.
Note that the uncertainty predictions vary for different voxel $i$.

\noindent\textbf{Data Uncertainty Learning (DUL)}
~\cite{chang2020data} shares a similar spirit with HAU with two distinctions. Instead of using the logit $\hat{\mathbf{z}}_i$, DUL models the feature $\hat{\mathbf{v}}_i$ as a Gaussian distribution by $\hat{\mathbf{v}}_i = \mathbf{v}_i + \bm{\sigma}_i^{\mathbf{W}} \epsilon $. Moreover, DUL introduces a regularization term in the loss function that  minimizes the 
Kullback–Leibler (KL) divergence between the predicted Gaussian and a standard Gaussian.

\noindent\textbf{MC Dropout (MCD)}
~\cite{jungo2019assessing} is proposed to explore the epistemic (model) uncertainty. Differently from the above methods, MCD does not require additional parameters to learn uncertainty. Instead, it incorporates multiple dropout layers into the original network during training. For inference, the occupancy prediction of each voxel is obtained by $\hat{\mathbf{z}}_i=\frac{1}{K} \sum_{k=1}^K \mathbf{z}_{k, i}$, where $\mathbf{z}_{k, i}$ is the model output at the ${k}$-th test. The normalized
entropy of $K$ predictions is adopted as the model uncertainty. To fully explore the uncertainty within the model, we set $K=40$ in our experiments.

For above online uncertainty learning methods, the calibrated confidence is set as the \textit{softmax} output of the sampled logit and then determined by taking the maximum probability across all classes: $c_i =  \max_s \mathcal{S}(\hat{\mathbf{z}}_i)^{(s)}$, where $\mathcal{S}$ denotes the \textit{softmax} function.

\subsection{Offline Model Calibration}
\label{sec:3model-calib}
Offline calibration methods build on pre-trained $V_{\theta}$ and need to learn a scaling function, which typically employ the following formulation
\begin{equation}
    c_i \equiv c_{f|\phi}(\mathbf{z}_i) = \max_s  f_\phi({\mathbf{z}_i})^{(s)} ~,
\end{equation}
where $f_\phi$ is the scaling function applied to $\mathbf{z}_i$ parameterized by the learnable parameters $\phi$. In the absence of explicit uncertainty estimation, uncertainty $\sigma_i$ is set to $1-c_i$.

\noindent\textbf{Temperature Scaling (TempS)~\cite{guo2017calibration}} employs a scalar parameter $T$, termed as temperature, to scale the logits $\mathbf{z_i}$, by $f_\phi(\mathbf{z_i}) = \mathcal{S}(\frac{\mathbf{z_i}}{T})$. $T$ is data-independent, which is shared across all classes.

\noindent\textbf{Dirichlet Scaling (DiriS)~\cite{kull2019dirichlet} }
 assumes that the model's output follow a Dirichlet distribution. Based on this assumption, they propose the Dirichlet scaling, $f_\phi(\mathbf{z_i}) = \mathcal{S}( \mathbf{W} \cdot \log (\mathcal{S}(\mathbf{z_i})) + \mathbf{b})$. Here, learnable parameters $\phi$ includes weight $\mathbf{W} \in \mathbb{R}^{(S+1)\times(S+1)}$ and bias $\mathbf{b} \in \mathbb{R}^{S+1}$.

\noindent\textbf{Meta-Calibration (MetaC)~\cite{ma2021metaCal}} proposes to use the entropy of model prediction  $-c_i\log(c_i)$ to select different calibrators.
Specifically, an identical $f_\phi(\mathbf{z_i}) = \mathcal{S}(\frac{\mathbf{z_i}}{T})$ as in TempS is used when $-c_i\log(c_i)$ is smaller than the predefined threshold $\eta$. Otherwise, the calibration function $f_\phi(\mathbf{z_i})$ is set to the constant value $\frac{1}{S+1}$.
MetaC introduces new randomness into predictions, which leads to variations in accuracy, making it less practical for safety-critical tasks such as occupancy prediction.

\noindent\textbf{Depth-Aware Scaling (DeptS)~\cite{kong2024calib3d} } is an improved variant upon MetaC, which is specially designed for LiDAR segmentation. Depth $d_i$ of each point or voxel is encoded into the calibration function $f_{\phi}$ by a linear mapping $\alpha_i = k_1\cdot d_i + k_2$, where $k_1$ and $k_2$ are learnable parameters.
When prediction entropy  $-c_i\log(c_i)$ is greater than the threshold $\eta$, $f_\phi(\mathbf{z_i}) = \mathcal{S}(\frac{\mathbf{z_i}}{\alpha \cdot T_1})$. Otherwise, $f_\phi(\mathbf{z_i}) = \mathcal{S}(\frac{\mathbf{z_i}}{\alpha \cdot T_2})$
Both $T_1$ and $T_2$ are temperature parameters, where $T_1$ is initially set higher than $T_2$.

\section{\model}
\label{sec:reliocc}
 
\begin{figure}[t]
\begin{center}
\includegraphics[width=1.0 \linewidth]{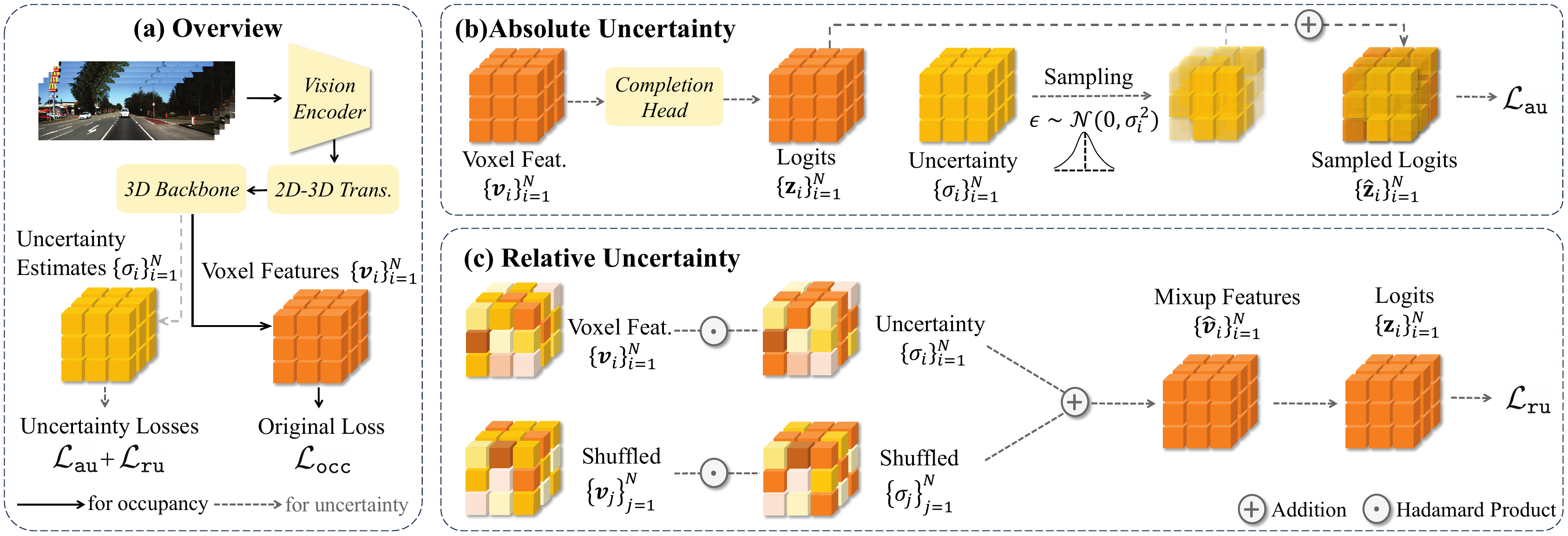} 
\end{center}
\vspace{-3mm}
\caption{(a) \emph{Overview of proposed \model}. 
Besides the original objective of an occupancy network, we introduce an uncertainty estimation branch and supervise it with absolute and relative uncertainty learning losses.
(b) \emph{Absolute uncertainty learning}. Deterministic logits are replaced with ones sampled from predicted distributions.
(c) \emph{Relative uncertainty learning}. We leverage the relative relationships between uncertainty pairs to further enhance uncertainty learning.
}
\vspace{-4mm}
 \label{fig:overallnetwork}
 \end{figure}

\noindent\textbf{Method Overview.} 
 Inspired by investigation on prior arts, we propose \model, a plug-and-play method tailored for the 3D occupancy prediction task. 
 \model has two main improvements over existing methods. 
 Firstly, beyond traditional uncertainty learning, we propose to utilize the relative relationships between voxel pairs of uncertainty estimates in order to further refine the uncertainty estimation process.
 Secondly, we present a unified framework that integrates the methodologies of uncertainty learning with scaling-based calibration, which demonstrates that their synergy offers substantial benefits. 
As shown in Figure~\ref{fig:overallnetwork}(a),  we predict a scalar uncertainty $\sigma_i$ from a voxel feature $\bm{v_i}$ using an MLP, which receives supervision from both the individual and relative voxels.

\noindent\textbf{Absolute Uncertainty.}
By \emph{absolute} here we mean the uncertainty $\sigma_i$ is only determined by the individual $\bm{v}_i$ itself while ignoring relative relations between a pair of $\bm{v}_i$ and $\bm{v}_j$. 
For absolute uncertainty, we adopt a similar formulation as in HAU~\cite{kendall2017uncertainties}.
Specifically, We randomly sample the logit $\hat{\mathbf{z}}_i$ based on the predicted uncertainty $\sigma_i$ by $\hat{\mathbf{z}}_i = \mathbf{z}_i + \sigma_i \epsilon, \epsilon \in \mathcal{N}(\mathbf{0}, \mathbf{I})$ as illustrated in Figure~\ref{fig:overallnetwork}(b). 
We denote this absolute uncertainty loss as $\mathcal{L}_\texttt{au}$, which is computed with the re-sampled logit $\hat{\mathbf{z}}_i$ and its corresponding ground truth.

\noindent\textbf{Relative Uncertainty.}
A potential drawback of absolute uncertainty is that the optimization of $\sigma_i$ tends to plateau once it reaches a small scale. To address this issue, we introduce the concept of relative uncertainty learning for occupancy prediction. The fundamental principle of relative uncertainty learning involves enforcing comparisons between uncertainty pairs $\bm{v}_i$ and $\bm{v}_j$. This approach ensures that optimization does not plateau, even when $\sigma_i$ and $\sigma_j$ are small.

Concretely, we shuffle voxel features in $\mathcal{V}$, paired the shuffled features with the original ones and obtain random pairs $(\bm{v}_i, \bm{v}_j)$ at each iteration. Inspired by the mix-up~\cite{zhang2018mixup} learning principle, we blend the paired voxel features with $\hat{\bm{v}} = \lambda\bm{v}_i + (1-\lambda)\bm{v}_j$. Correspondingly $\hat{\bm{v}}$ is trained with a blend of the label pairs using cross-entropy loss. The blended label $y = y_i + y_j$, where $y_i,y_j$ are one-hot label encodings\footnote{Different from the original mix-up~\cite{zhang2018mixup} paper, we omit weighting the labels with $\lambda$ for stable training.}. We employ the predicted uncertainty for the weighting: $\lambda = \frac{\sigma_i}{\sigma_i + \sigma_j} \in [0,1]$. We denote the loss computed with blended label and mixed output from the shared completion head as the relative uncertainty loss $\mathcal{L}_\texttt{ru}$, as shown in Figure~\ref{fig:overallnetwork}(c).

An intuitive understanding emerges when considering that relative uncertainty adaptively modulates the learning dynamics between the feature pair $(\bm{v}_i, \bm{v}_j)$. Specifically, if the model exhibits greater confidence in the prediction associated with $\bm{v}_i$, it suffices for the mixup feature to incorporate a smaller portion  of  $\bm{v}_i$ while still achieving a reduced loss $\mathcal{L}_\texttt{ru}$. In contrast, a lower confidence in $\bm{v}_j$ necessitates a greater inclusion of $\bm{v}_j$ within the mixup to diminish the loss. Consequently, this process enables the model to effectively differentiate between the uncertainties $\sigma_i$ and $\sigma_j$, typically resulting in a smaller $\sigma_i$ and a larger $\sigma_j$. 
Importantly, this differentiation does not hinge on the absolute magnitudes of  $\sigma_{i,j}$. Rather, it is the relative relationship between them that is central to the learning process. 
In driving scenarios, there are rich relative relationships between voxels, including \textit{distance}, \textit{occupancy}, \textit{surface} and \textit{interior} properties.
This focus on relative differences ensures that the model's adjustments are robust to the absolute scales of the uncertainties.

\noindent\textbf{Uncertainty-Aware Calibration.}
\label{adpat-model-calib}
Using the above two uncertainty estimation objective, \model is capable of learning uncertainty with existing occupancy models in an online setting. 
We introduce a scaling-based calibration objective to make it also compatible with the offline setting. A variant form of TempS~\cite{guo2017calibration} is adopted, and the uncertainty-aware temperature $T_\sigma$ is a linear transform of $\sigma_i$: 
\begin{equation}
    T_\sigma = k_1\cdot \sigma_i + k_2,~~~~f_\phi(\mathbf{z_i}) = \mathcal{S}\left ( \frac{\mathbf{W} \cdot \mathbf{z_i} + \mathbf{b}}{T_\sigma} \right),
\end{equation}

where $k_1$ and $k_2$ are learnable parameters, and $\mathbf{b}$ is the bias. $\mathbf{W}$ is initialized as the identity matrix and only the elements on the diagonal are optimized. The calibration loss is denoted as $\mathcal{L}_\texttt{calib}$.

\noindent\textbf{Training and Inference.}
\model supports both online uncertainty learning and offline model calibration settings. In the online setting, the uncertainty predictor is trained concurrently with the occupancy network from scratch. The total loss function comprises $\mathcal{L}_\texttt{occ}$, $\mathcal{L}_\texttt{au}$, and $\mathcal{L}_\texttt{ru}$. Here, $\mathcal{L}_\texttt{occ}$ represents the primary loss for the occupancy network. During inference, the model operates consistently with the original design as the predicted uncertainties are not utilized.
In the offline setting, the occupancy network is fixed, eliminating the need for $\mathcal{L}_\texttt{occ}$ and introducing the calibration loss $\mathcal{L}_\texttt{calib}$ instead. The inference process incurs a minimal increase in computational overhead due to the addition of the calibrator.

\section{Experiments}
\label{sec:exp}
\textbf{Datasets and Evaluation.}
SemanticKITTI~\cite{behley2019semantickitti} is the first large-scale outdoor dataset for semantic occupancy prediction containing 64-beam LiDAR point clouds and camera images as inputs~\cite{geiger2012we}.
The dataset comprises 22 sequences, where 00-10 (excluding 08) are used as the training set, 08 is the validation set, and 11-21 are the test set. 
Since the ground truth for the test set is not publicly available, we cannot measure our newly introduced metrics on it. Therefore, we primarily evaluate existing methods on the validation set (\texttt{val.}). As described in \S\ref{eval-metric}, mIoU and IoU are used to measure the model's accuracy in semantic and geometric completion, respectively. 
For misclassification detection and calibration metrics including \texttt{PRR} and \texttt{ECE}, we also report the corresponding results from both geometric and semantic perspectives.

\newcommand{\reshl}[2]{
\textbf{#1} \fontsize{7.5pt}{1em}\selectfont\color{mygreen}{$\!\uparrow\!$ \textbf{#2}}
}
\setlength\intextsep{0pt}

\begin{table}[t]
\setlength{\abovecaptionskip}{0cm}
\begin{center}
    \caption{The accuracy and reliability evaluation of state-of-the-art semantic occupancy prediction models on the ~\texttt{validation set} of SemanticKITTI~\cite{behley2019semantickitti}. $*$ indicates that the output of SCPNet is a sparse representation and does not contain confidence score for empty voxels, making it infeasible to evaluate the corresponding geometric metrics in reliability.}
    \vspace{0.1em}
    \label{tab:benchmark}
    {\hspace{-1.3ex}
    \resizebox{0.95\textwidth}{!}{
    \setlength\tabcolsep{2.2mm}
    \begin{tabular}{r||c||ccc||ccc}
      \hline\thickhline
      \rowcolor{mygray}
      &   & \multicolumn{3}{c||}{\textbf{Semantics} } & \multicolumn{3}{c}{\textbf{Geometry}} \\
      \rowcolor{mygray}
       \multicolumn{1}{c||}{\multirow{-2}{*}{Method}}
      & \multicolumn{1}{c||}{\multirow{-2}{*}{Modality}}  &
        \multicolumn{1}{c}{{mIoU (\%)$\uparrow$}} & \multicolumn{1}{c}{{$\texttt{PRR}_{\texttt{sem}}$(\%)$\uparrow$}} & \multicolumn{1}{c||}{{$\texttt{ECE}_{\texttt{sem}}$(\%)$\downarrow$}} &  \multicolumn{1}{c}{{IoU(\%)$\uparrow$}}  & \multicolumn{1}{c}{{$\texttt{PRR}_{\texttt{geo}}$(\%)$\uparrow$}} & \multicolumn{1}{c}{{$\texttt{ECE}_{\texttt{geo}}$(\%)$\downarrow$}} \\\hline\hline
      {SSCNet~\pub{CVPR17}}{~\cite{song2017semantic}}                 &  {LiDAR}  & 16.41 & 46.77 & 1.61  &  50.75 & 42.92 & 0.97  \\
      {LMSCNet~\pub{3DV20}}{~\cite{roldao2020lmscnet}}                 &  {LiDAR}  & 17.27 & \textbf{48.89} & \textbf{0.79} & \textbf{54.91} & \textbf{48.01}  & \textbf{0.67}  \\
      {JS3C-Net~\pub{AAAI21}}{~\cite{song2017semantic}}                 &  {LiDAR}  & 22.77 & 41.09 & 2.94 & 53.08  & 37.04 & 1.64  \\
      {SSC-RS~\pub{IROS23}}{~\cite{mei2023ssc}}                 &  {LiDAR}  & 24.75 & 45.04 & 0.87 & 58.62 & 44.29  & 0.72  \\
      {SCPNet$^*$~\pub{CVPR23}}{~\cite{xia2023scpnet}}                 &  {LiDAR}  & \textbf{35.06} & 38.35 & 2.52  & 49.06 & - & -   \\
      \cdashline{1-8}[1pt/1pt]
      {MonoScene~\pub{CVPR22}}{~\cite{cao2022monoscene}}           & {Camera}  & 11.30 & 41.95 & 6.65 & 36.79 & 38.39 & 5.95\\
      TPVFormer~\pub{CVPR23}{~\cite{huang2023tri}} & {Camera}  & 11.30 & 38.83 & 7.10 & 35.62 & 32.10 & 6.32  \\ 
      NDCScene~\pub{ICCV23}{~\cite{yao2023ndc}} & {Camera}  & 12.70 & 43.29 & 7.24 & 37.24 & \textbf{40.17} & 6.45  \\ 
      VoxFormer~\pub{CVPR23}{~\cite{li2023voxformer}} & {Camera}  & 13.17 & 42.97  & 5.90  &  43.96 &  36.56 & 5.02   \\ 
      SGN~\pub{arXiv23}{~\cite{mei2023camera}} & {Camera}  & \textbf{15.52} & \textbf{44.72}  & \textbf{5.69}  & \textbf{45.45}  & 39.78  &  \textbf{4.85}  \\ 
      \hline
    \end{tabular}
    }
    }
\end{center}
\vspace{-16pt}
\end{table}

\textbf{Re-evaluated Methods.} We reproduce and evaluate existing publicly available methods on the SemanticKITTI benchmark, including five LiDAR-based models~\cite{song2017semantic,roldao2020lmscnet,yan2021sparse,mei2023ssc,xia2023scpnet} and five camera-based models~\cite{cao2022monoscene,huang2023tri,li2023voxformer,yao2023ndc,mei2023camera}. 
All results are obtained using the official implementation and the configurations are kept consistent for inference, with relevant links provided in our supplemental material.
As shown in Tab.~\ref{tab:benchmark}, we find that although the accuracy of camera-based methods has been continuously improved and gradually approaches the baseline accuracy of LiDAR methods, their reliability metrics, particularly the \texttt{ECE}, have not shown corresponding improvements.
In cases of lower accuracy compared to LiDAR, the camera-based models' reliability is also quite poor, which undoubtedly poses significant safety risks for autonomous driving.

\begin{table}[t]
\setlength{\abovecaptionskip}{0cm}
\begin{center}
    \caption{Quantitative results of online uncertainty learning (\S\ref{exp:unc-learning}) on SemanticKITTI~\cite{behley2019semantickitti} (\texttt{val.}).}
    \vspace{0.1em}
    \label{tab:uncsota}
    {\hspace{-1.3ex}
    \resizebox{0.95\textwidth}{!}{
    \setlength\tabcolsep{2.2mm}
    \begin{tabular}{l||ccc||ccc}
      \hline\thickhline
      \rowcolor{mygray}
      & \multicolumn{3}{c||}{\textbf{Semantics} } & \multicolumn{3}{c}{\textbf{Geometry}} \\
      \rowcolor{mygray}
       \multicolumn{1}{c||}{\multirow{-2}{*}{Method}} &
        \multicolumn{1}{c}{{mIoU (\%)$\uparrow$}} & \multicolumn{1}{c}{{$\texttt{PRR}_{\texttt{sem}}$(\%)$\uparrow$}} & \multicolumn{1}{c||}{{$\texttt{ECE}_{\texttt{sem}}$(\%)$\downarrow$}} &  \multicolumn{1}{c}{{IoU(\%)$\uparrow$}}  & \multicolumn{1}{c}{{$\texttt{PRR}_{\texttt{geo}}$(\%)$\uparrow$}} & \multicolumn{1}{c}{{$\texttt{ECE}_{\texttt{geo}}$(\%)$\downarrow$}} \\\hline
      \hline
      \rowcolor{mygray1}
      \multicolumn{7}{c}{\small \em VoxFormer Framework} \\
      \hline
      VoxFormer~\pub{CVPR23}{~\cite{li2023voxformer}}   & 13.17 & 42.97  & 5.90  &  43.96 &  36.56 & 5.02   \\ 
      VoxFormer+HAU~\pub{NIPS17}{~\cite{kendall2017uncertainties}}   & \textbf{13.43} & 45.38  &  5.26 & 43.57  &  40.72 & 4.47   \\ 
      VoxFormer+DUL~\pub{CVPR20}{~\cite{chang2020data}}  & 13.29  &  43.57 & 6.09  & \textbf{44.10}  &  38.66 & 5.17   \\ 
      VoxFormer+MCD~\pub{MICCAI19}{~\cite{jungo2019assessing}}   & 13.28 & 42.21  & 5.83  & 43.90  &  37.43 & 4.99   \\ 
      VoxFormer+\model({Ours})      & \textbf{13.43} & \textbf{47.75} & \textbf{2.84}  & 43.28   & \textbf{44.58} &  \textbf{2.57} \\
      \hline
      \rowcolor{mygray1}
      \multicolumn{7}{c}{\small \em SGN Framework} \\
       \hline
      SGN~\pub{arXiv23}{~\cite{mei2023camera}}   & 15.52 & 44.72  & 5.69  & 45.45  & 39.78  &  4.85  \\ 
      SGN+HAU~\pub{NIPS17}{~\cite{kendall2017uncertainties}}   & 15.50  & 46.51  &  5.08 & 45.07  & 44.24   &  4.34  \\ 
      SGN+DUL~\pub{CVPR20}{~\cite{chang2020data}}   & \textbf{15.81} & 44.00  &  5.78 &  45.75 &  39.56 &  4.95  \\ 
      SGN+MCD~\pub{MICCAI19}{~\cite{jungo2019assessing}}   & 15.62 &  44.70 &  6.02 &  45.50 &  40.34 &  5.11  \\ 
      SGN+\model({Ours})                  & 15.65  & \textbf{50.72}  & \textbf{3.75} &  \textbf{45.78} & \textbf{49.61}  &  \textbf{3.07} \\
      \hline
    \end{tabular}
    }
    }
\end{center}
\vspace{-5mm}
\end{table}

 \begin{figure}[t]
\begin{center}
\includegraphics[width=1.0\linewidth]{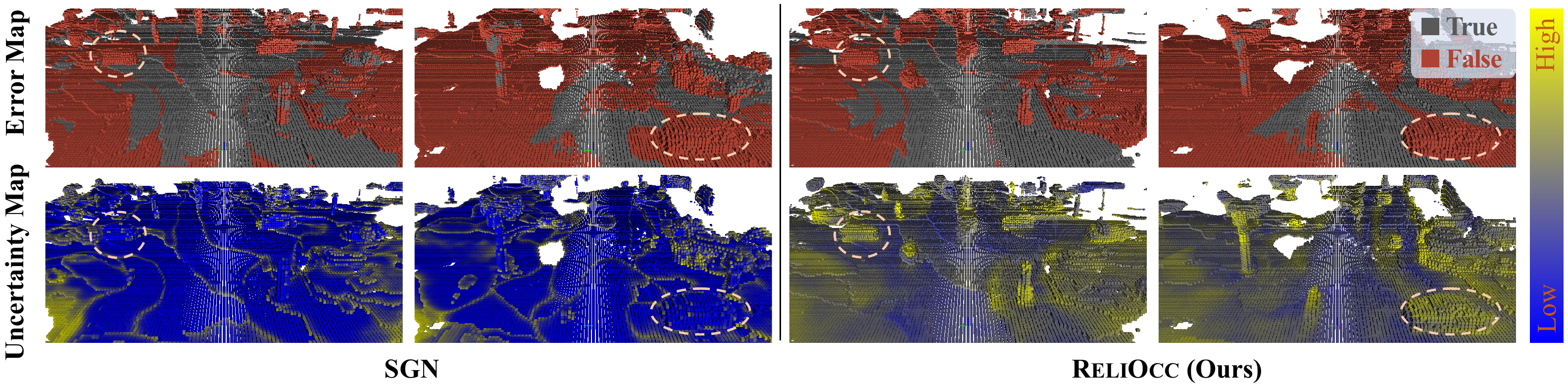} 
\end{center}
\vspace{-3mm}
\caption{Visual results of the error map and uncertainty map from the prediction by SGN~\cite{mei2023camera} and \model.
In the uncertainty map, a closer proximity to \textit{yellow} indicates a higher level of uncertainty.}
 \label{fig:unc}
 \vspace{-5mm}
 \end{figure}

\subsection{Online Uncertainty Learning}
\label{exp:unc-learning}

\textbf{Base Architectures and Competing Methods.}
Considering the potential applications of camera-based methods and their current limitations, we adopt the state-of-the-art vision-based methods including VoxFormer~\cite{li2023voxformer} and SGN~\cite{mei2023camera} as our base architectures to conduct relevant experiments. 
For the online uncertainty learning, we report the results of some existing methods on the two baseline frameworks for comparison, including HAU~\cite{kendall2017uncertainties}, DUL~\cite{chang2020data}, and MCD~\cite{jungo2019assessing} (see \S\ref{sec:3unc-learn}). 

\textbf{Implementation Details.}
We follow the original training setting and add additional uncertainty learning parameters without altering the network structure. The inputs consist of the current image from the left camera and previous $4$ frames. The image size is copped into $1220 \times 370$. VoxFormer~\cite{li2023voxformer} and SGN~\cite{mei2023camera} with online uncertainty estimation are trained for $20$ epochs and $40$ epochs, respectively. 
The loss coefficients ($\alpha, \beta$) for $\mathcal{L}_\texttt{au}$ and $\mathcal{L}_\texttt{ru}$ are set to ($4.0$, $6.0$) for both frameworks. All the experiments are conducted on 8 NVIDIA Tesla V100 GPUs within the same experimental environment. Further implementation details can be found in the supplemental material.

\textbf{Quantitative Results.} 
In Tab.~\ref{tab:uncsota}, we provide the comparison among the methods with the same framework for fairness.
Compared to data uncertainty-based HAU~\cite{kendall2017uncertainties} and DUL~\cite{chang2020data}, as well as model uncertainty-based MCD~\cite{jungo2019assessing}, our method shows significant improvements in the new evaluation metrics for reliability. The calibration errors (\texttt{ECE}) in both semantic and geometric aspects are significantly reduced compared to the existing uncertainty estimation methods. The improvement in \texttt{PRR} also indicates a notable enhancement in model reliability. 
Our method maintains stability in terms of the original accuracy (mIoU and IoU) across the two different frameworks.

\textbf{Qualitative Results.} 
To further verify the effect of our method, we visualize the error map with corresponding uncertainty map of SGN~\cite{mei2023camera} and our approach.
The uncertainty for vanilla SGN is obtained by subtracting the confidence from $1$. As shown in Figure ~\ref{fig:unc}, when the network's predictions exhibit large areas of error, SGN's uncertainty map still shows low uncertainty, indicating high confidence in prediction. In contrast, our proposed \model displays high uncertainty in most of the error regions, providing more reliable information for downstream tasks.

\subsection{Offline Model Calibration}
\label{exp:model-calib}

In this section, VoxFormer and SGN are also adopted as baseline frameworks. We primarily compare our method with scaling-based model calibration approaches including TempS~\cite{guo2017calibration}, DiriS~\cite{kull2019dirichlet}, MetaC~\cite{ma2021metaCal}, and DeptS~\cite{kong2024calib3d} (see \S\ref{sec:3model-calib}).

\textbf{Implementation Details.}
We select the best-performing model checkpoints on the validation set from the pre-trained VoxFormer and SGN as the targets for calibration.  During the calibration process, the parameters of the original network are frozen, and only the parameters $\phi$ in the calibration function $f_\phi$ and uncertainty learning layers are trainable. For both frameworks, these methods are trained on 8 GPUs for $20$ epochs with a learning rate as $0.001$ and AdamW optimizer~\cite{zhao2022decoupled}. The batch size is set to $1$ per GPU. For our method, the loss weights $(\alpha, \beta, \gamma)$ for uncertainty learning ($\mathcal{L}_\texttt{au}$, $\mathcal{L}_\texttt{ru}$) and model calibration ($\mathcal{L}_\texttt{calib}$) are set to $1.5$, $1.0$, and $4.0$, respectively.

\textbf{Quantitative Results.} 
As illustrated in Tab.~\ref{tab:calsota}, all model calibration methods demonstrate improvements compared to the baselines, particularly in calibration error (\texttt{ECE}).
MetaC~\cite{ma2021metaCal} loses  the characteristic of maintaining accuracy in calibration due to the introduction of new random classifications.
Our approach with uncertainty-aware design achieves competitive performance on both \texttt{ECE} and \texttt{PRR} metrics without depth information even compared with the state-of-the-art DeptS~\cite{kong2024calib3d}.

\begin{table}[t]
\setlength{\abovecaptionskip}{0cm}
\begin{center}
    \caption{Quantitative results of offline model calibration (\S\ref{exp:model-calib}) on SemanticKITTI~\cite{behley2019semantickitti} (\texttt{val.}).}
    \vspace{0.1em}
    \label{tab:calsota}
    {\hspace{-1.3ex}
    \resizebox{0.95\textwidth}{!}{
    \setlength\tabcolsep{2.2mm}
    \begin{tabular}{l||ccc||ccc}
      \hline\thickhline
      \rowcolor{mygray}
        & \multicolumn{3}{c||}{\textbf{Semantics} } & \multicolumn{3}{c}{\textbf{Geometry}} \\
      \rowcolor{mygray}
       \multicolumn{1}{c||}{\multirow{-2}{*}{Method}} &
        \multicolumn{1}{c}{{mIoU(\%)$\uparrow$}} & \multicolumn{1}{c}{{$\texttt{PRR}_{\texttt{sem}}$(\%)$\uparrow$}} & \multicolumn{1}{c||}{{$\texttt{ECE}_{\texttt{sem}}$(\%)$\downarrow$}} &  \multicolumn{1}{c}{{IoU(\%)$\uparrow$}}  & \multicolumn{1}{c}{{$\texttt{PRR}_{\texttt{geo}}$(\%)$\uparrow$}} & \multicolumn{1}{c}{{$\texttt{ECE}_{\texttt{geo}}$(\%)$\downarrow$}} \\\hline\hline
      \rowcolor{mygray1}
      \multicolumn{7}{c}{\small \em VoxFormer Framework} \\
      \hline
      VoxFormer~\pub{CVPR23}{~\cite{li2023voxformer}}   & 13.17 & 42.97  & 5.90  &  43.96 &  36.56 & 5.02\\ 
      VoxFormer+TempS~\pub{ICML17}{~\cite{guo2017calibration}}   & 13.17 &  43.63 & 2.61  & 43.96  &  33.59 & 2.28 \\
      VoxFormer+DiriS~\pub{NeurIPS19}{~\cite{kull2019dirichlet}}    & 13.17 & 48.12  & 2.38  & 43.96  &  42.78 &  2.42  \\ 
      VoxFormer+MetaC~\pub{ICML21}{~\cite{ma2021metaCal}}    & 11.86 & 43.06  &  4.11 & 34.73  & 34.80  &  3.67  \\ 
      VoxFormer+DeptS~\pub{arXiv24}{~\cite{kong2024calib3d}}    & 13.17 & 41.29  & 2.27 & 43.96 & 30.31  &  \textbf{1.63}  \\ 
      VoxFormer+\model({Ours})       & 13.17  &  \textbf{48.17} & \textbf{2.05} &  43.96 & \textbf{44.34} & 2.57  \\
      \hline
      \rowcolor{mygray1}
      \multicolumn{7}{c}{\small \em SGN Framework} \\
       \hline
      SGN~\pub{arXiv23}{~\cite{mei2023camera}}  & 15.52 & 44.72  & 5.69  & 45.45  & 39.78  &  4.85  \\ 
      SGN+TempS~\pub{ICML17}{~\cite{guo2017calibration}}   & 15.52 & 46.90  & 2.68  & 45.45  &  37.25 &  2.35  \\ 
      SGN+DiriS~\pub{NeurIPS19}{~\cite{kull2019dirichlet}}    & 15.52 & \textbf{48.20}  & 2.61  & 45.45  & 43.04  & 2.51   \\ 
      SGN+MetaC~\pub{ICML21}{~\cite{ma2021metaCal}}  & 14.71 & 46.38  & 4.06  &  40.37 & 37.97   &  3.65  \\ 
      SGN+DeptS~\pub{arXiv24}{~\cite{chang2020data}}    & 15.52 &  45.45 & 2.14 &  45.45 & 34.99  &  \textbf{1.42}  \\ 
      SGN+\model({Ours})   & 15.52 & 47.40  & \textbf{2.09}  &  45.45 & \textbf{43.80}  & 2.43  \\
      \hline
    \end{tabular}
    }
    }
\end{center}
\vspace{-6mm}
\end{table}

\begin{wrapfigure}{r}{0.5\linewidth}
\centering
\includegraphics[width=0.5\textwidth]{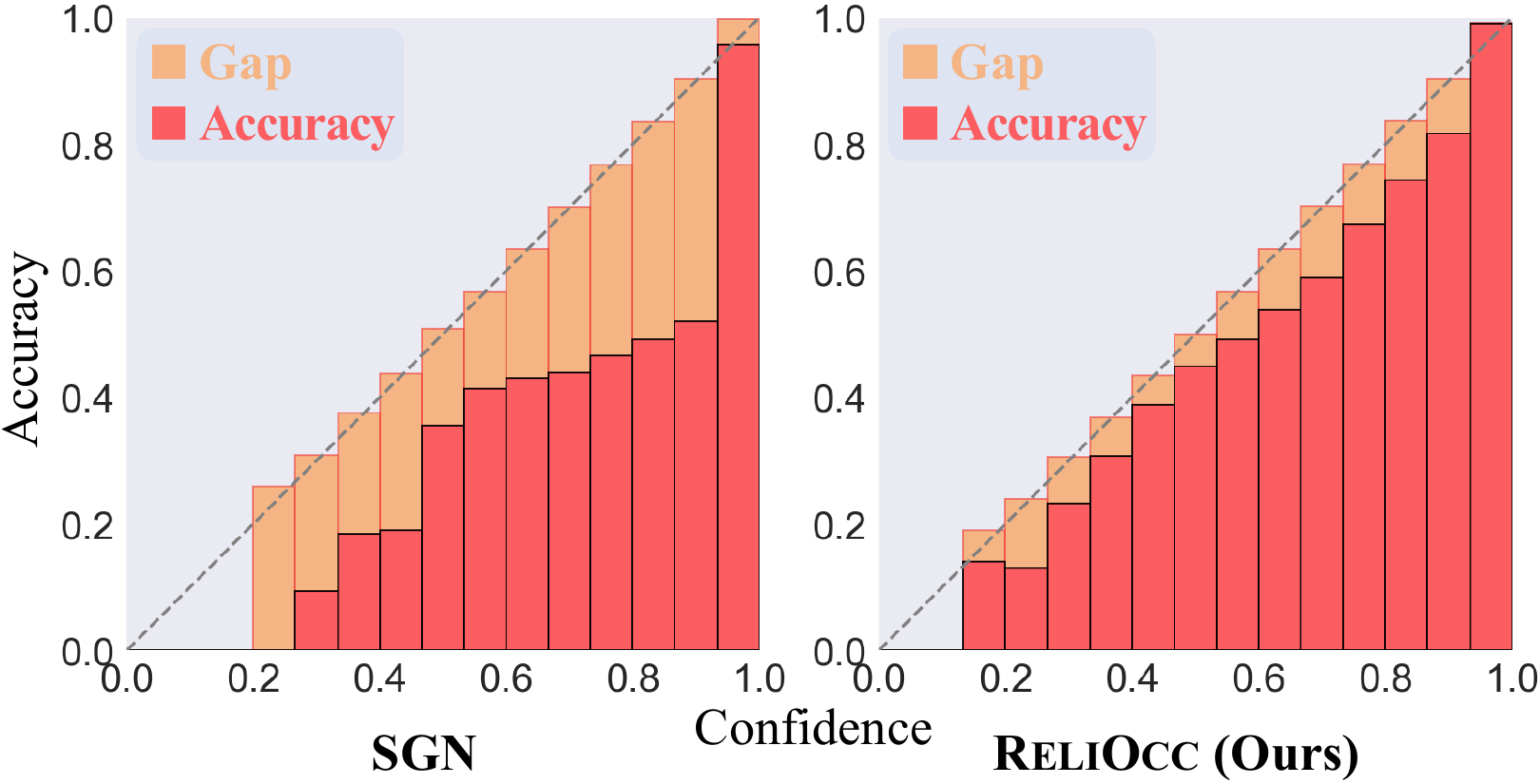}
\caption{The reliability diagrams of semantic calibration gaps from SGN~\cite{mei2023camera} and \model.}
\label{fig:reli_diag}
\end{wrapfigure}
\textbf{Qualitative Results.} 
We visualize the semantic calibration gaps in SGN~\cite{mei2023camera} and \model by reliability diagrams~\cite{degroot1983comparison, niculescu2005predicting}, which plot the confidence of prediction with the corresponding voxel-wise accuracy. 
Our method shows significant improvement over the baseline model in calibration gaps, as indicated by the Gap areas between the predicted accuracy with the expected accuracy (smaller indicates better), as shown in the Figure~\ref{fig:reli_diag}. Additional reliability diagrams are provided in the supplemental material.

\subsection{Diagnostic Experiments}
\label{sec:robust_abl}
To further validate the effectiveness of our approach, we present diagnostic experiments in this section. Due to page limitations, some experimental results are provided in the supplemental material.

\begin{wraptable}{r}{0.52\linewidth}
	\centering
	\setlength{\abovecaptionskip}{0cm}
    \captionsetup{width=.52\textwidth}
    \caption{Ablation of our online uncertainty learning.}
    \label{tab:abl_online}
    {
    \resizebox{0.52\textwidth}{!}{
    \setlength\tabcolsep{4pt}
    \renewcommand\arraystretch{1.0}
    \begin{tabular}{cc||cc|cc}
        \hline\thickhline
        \rowcolor{mygray}
            Absolute Unc. & Relative Unc. & $\texttt{PRR}_{\texttt{sem}}$$\uparrow$ & $\texttt{ECE}_{\texttt{sem}}$$\downarrow$ & $\texttt{PRR}_{\texttt{geo}}$$\uparrow$ & $\texttt{ECE}_{\texttt{geo}}$$\downarrow$  \\ 
        \hline\hline
          &  &   42.97 & 5.90 & 36.56 & 5.02  \\
           \cmark & & 45.47  & 5.41 & 41.54 &  4.62 \\
               &\cmark & 46.58   & 4.02 & 42.65 & 4.31   \\
           \cmark     &\cmark &  \textbf{47.75}  & \textbf{2.84} & \textbf{44.58} & \textbf{2.57}  \\
        \hline
    \end{tabular}
        }
    }
\end{wraptable}

\textbf{Ablation of Online Uncertainty Learning.} 
We provide ablation experiments on the effect of absolute uncertainty and relative uncertainty during the whole model training. The experiments are conducted  with VoxFormer~\cite{li2023voxformer} on the validation set of SemanticKITTI.
As shown in Tab.~\ref{tab:abl_online}, the first row presents the baseline results. The inclusion of individual absolute uncertainty and relative uncertainty both contribute to the improvement of the model's reliability, although the improvement is relatively limited. 
When our proposed hybrid uncertainty learning module is incorporated, the  \texttt{PRR} and \texttt{ECE} metrics of model's prediction achieve the best results.

\begin{wraptable}{r}{0.62\linewidth}
	\centering
 \vspace{-0.2em}
	\setlength{\abovecaptionskip}{0cm}
    \captionsetup{width=.62\textwidth}
    \caption{Ablation of our offline model calibration.}
    \label{tab:abl_offline}
    {\hspace{-1.5ex}
    \resizebox{0.62\textwidth}{!}{
    \setlength\tabcolsep{4pt}
    \renewcommand\arraystretch{1.0}
    \begin{tabular}{ccc||cc|cc}
        \hline\thickhline
        \rowcolor{mygray}
           Scaling Calib. & Relative Unc. &  Absolute Unc. & $\texttt{PRR}_{\texttt{sem}}$$\uparrow$ & $\texttt{ECE}_{\texttt{sem}}$$\downarrow$ & $\texttt{PRR}_{\texttt{geo}}$$\uparrow$ & $\texttt{ECE}_{\texttt{geo}}$$\downarrow$  \\ 
        \hline\hline
         &  &  &   42.97 & 5.90 & 36.56 & 5.02   \\
        \cmark &  & & 43.63 & 2.61  & 33.59 &  2.28 \\
        \cmark  &   \cmark    & &  45.13  & \textbf{1.75} & 39.66 & \textbf{2.19}   \\
         \cmark  &  \cmark     &\cmark & \textbf{48.17 }& 2.05  &  \textbf{44.34} &  2.57  \\
        \hline
    \end{tabular}
        }
    }
\end{wraptable}
\textbf{Ablation of Offline Model Calibration.} 
Further ablations are also conducted in offline mode. 
With the pre-trained VoxFormer, we found that employing standard scaling strategies such as TempS~\cite{guo2017calibration} can achieve good calibration results as illustrated in second row of Tab.~\ref{tab:abl_offline}. 
However, it impacts the improvement of misclassification detection metrics (\texttt{PRR}) and even leads to a decline in geometry.
Our introduced relative uncertainty learning can further improve calibration performance and enhance misclassification detection. Furthermore, the combination of absolute and relative uncertainties achieves the best performance in misclassification detection, although it is slightly less effective in calibration compared to using relative uncertainty alone.

\begin{figure}[t]
\begin{center}
\includegraphics[width=1.0\linewidth]{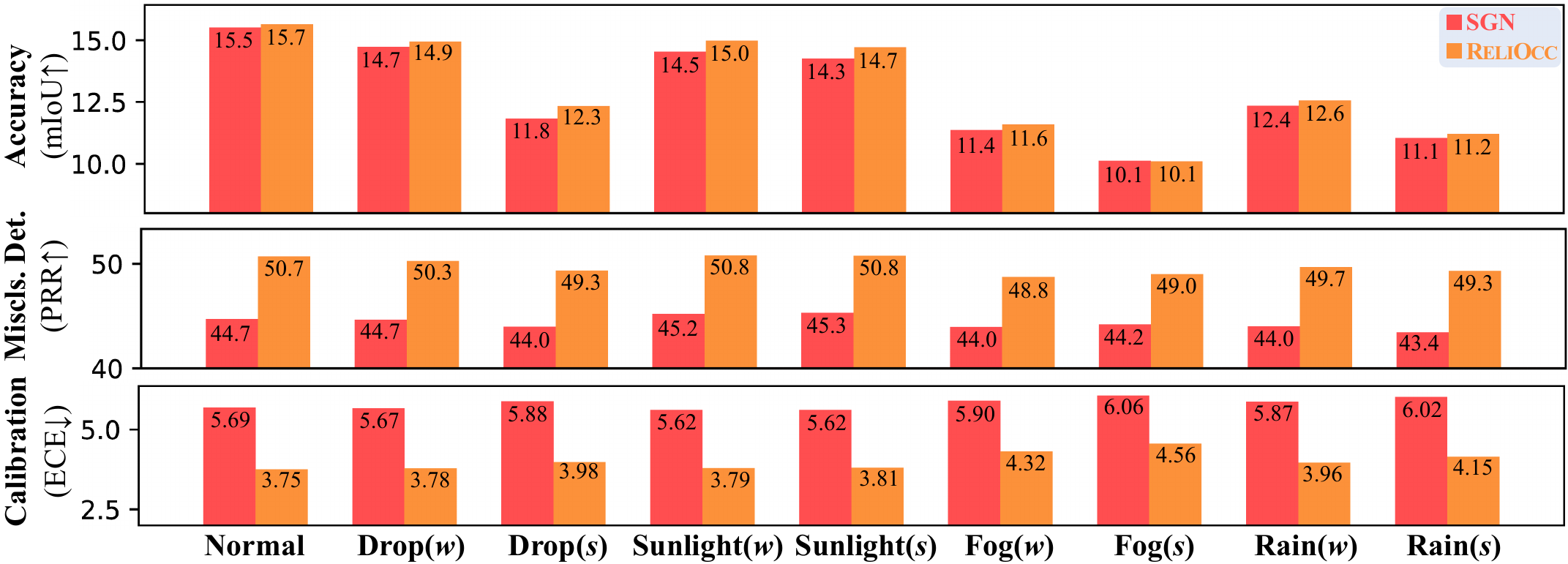} 
\end{center}
\vspace{-3mm}
\caption{The comparison in accuracy and reliability performance between SGN~\cite{mei2023camera} and \model under four out-of-domain conditions. Due to space constraints, we provide the semantic comparison here, while the geometric comparison is included in the supplemental material.}
\vspace{-5.5mm}
 \label{fig:robust}
 \end{figure}

\textbf{Robustness Analysis.}
Fig~\ref{fig:robust} presents the robustness analysis results of \model compared to the baseline model SGN~\cite{mei2023camera}. 
We simulate four potential out-of-domain scenarios during the inference, including sensor failures (frames drop), strong sunlight, foggy and rainy conditions, to evaluate the model's robustness~\cite{dong2023benchmarking}.
Each adverse scenario provides \textit{weak}(\textit{w}) and \textit{strong}(\textit{s}) modes of perturbation.
As the noise increases in various conditions, our method not only maintains stability in reliability metrics but also demonstrates more improvement in accuracy compared to the baseline.

\section{Related Work}
\label{sec:rlt-work}
\noindent \textbf{Semantic Occupancy Prediction.}
Semantic occupancy prediction (SOP) is also known as semantic scene completion (SSC) and firstly explored in indoor scenes~\cite{song2017semantic, liu2018see, li2020attention, cai2021semantic, tang2022not}. 
In outdoor scenarios, SemanticKITTI~\cite{behley2019semantickitti} stands as the first large-scale dataset, providing abundant data resources. 
Recently, several other datasets~\cite{wilson2022motionsc, wang2023openoccupancy, wei2023surroundocc,tian2024occ3d} have been constructed to explore this task owing to its importance. 
LiDAR-based methods~\cite{roldao2020lmscnet, rist2021semantic_implicit, yang2021semantic_assisted,zuo2023pointocc, cheng2021s3cnet, yan2021sparse, wilson2022motionsc, xia2023scpnet} have dominated this field in accuracy. 
MonoScene~\cite{cao2022monoscene} is the first occupancy prediction method that utilizes single image as input. Subsequent studies~\cite{huang2023tri, li2023voxformer, mei2023camera, lu2023octreeocc,jiang2023symphonize, yu2023flashocc,li2023fb_occ, wang2023panoocc, lyu2023_3dopformer, ma2023cam4docc, wang2024not, li2023stereoscene, wang2024label, shi2024occfiner,Tang_2024_CVPR,wang2024occgen,veon} have effectively improved the performance of camera-based models.  
However, there is a lack of research on reliability of occupancy predictions, posing potential risks to the safety of downstream tasks in driving~\cite{zheng2023occworld, hu2023planning, albrecht2021interpretable}. 
\model fills this gap by investigating the reliability of occupancy networks via uncertainty learning.

\noindent \textbf{Uncertainty Learning and Model Calibration.} 
The uncertainty in machine learning consists of aleatoric uncertainty from data noises and epistemic uncertainty from model parameters~\cite{kendall2017uncertainties, kendall2018multi}.
Data uncertainty is widely explored in face field~\cite{chang2020data, xu2014data, zhang2021relative, gong2017capacity, khan2019striking,relireid}.
Cai~\etal~\cite{cai2023uncertainty} propose a probabilistic embedding model to estimates the data uncertainty for point cloud.
Model uncertainty is usually obtained from the statistics of multiple predictions through methods including model ensembling~\cite{lakshminarayanan2017simple}, bootstrapping~\cite{johnson2001introduction}, and bagging~\cite{breiman1996bagging}.
Model calibration is another line to improving reliability in model prediction~\cite{de2023reliability}, which provides a post-processing scheme applied to the non-probabilistic output from a trained model. 
Model calibration was initially studied in image classification~\cite{guo2017calibration, vaicenavicius2019evaluating, kull2019dirichlet,ma2021metaCal} and has since been widely applied to 
object detection~\cite{kuppers2020multivariate, munir2022towards, oksuz2023towards,popordanoska2024beyond} and semantic segmentation~\cite{ding2021local, elias2021calibrated, wang2023selective, kong2024calib3d}.
Our method adopts uncertainty as a learning objective and can support both online uncertainty estimation and offline model calibration simultaneously.

\section{Conclusion}
\label{sec:conc}
In this paper, we address the issue of assessing reliability in semantic occupancy prediction for the first time. 
The reliability is evaluated from the two aspects including misclassification detection and calibration.
Exhaustive evaluation of existing LiDAR and camera-based methods are provided. Besides, we propose a new scheme \model that integrates hybrid uncertainty from the individual and relative voxels into existing occupancy networks without affecting accuracy or inference speed. 
Both online and offline modes are designed to illustrate the generalization capability of our learned uncertainty.
Extensive experiments are conducted under various settings, demonstrating \model is effective in improving the reliability and robustness of semantic occupancy models.

\medskip

{\small
\bibliographystyle{unsrt}
\bibliography{main}
}


\newpage
\appendix
\section*{Supplemental Material}

\setcounter{figure}{0}
\setcounter{table}{0}
\renewcommand{\thefigure}{A\arabic{figure}}
\renewcommand{\thetable}{A\arabic{table}}

In this part, we further provide more details, additional experimental results and discussions on our proposed approach:
\begin{itemize}[leftmargin=*]
	\setlength{\itemsep}{0pt}
	\setlength{\parsep}{-0pt}
	\setlength{\parskip}{-0pt}
	\setlength{\leftmargin}{-10pt}
	\vspace{-2pt}
  \item \S\ref{sec:supp_detail}: More implementation details;
  \item \S\ref{sec:supp_exp}: Additional experiments; 
  \item \S\ref{sec:supp_dis}: Further discussions;
  \item \S\ref{sec:supp_license}: License and consent with public resources.
\end{itemize}

\section{More Implementation Details}
\label{sec:supp_detail}

\subsection{Online Uncertainty Learning}
\textbf{VoxFormer Framework.}
VoxFormer~\cite{li2023voxformer} is a two-stage semantic occupancy prediction model with a highly efficient design. To simplify the problem, we use the results of the Stage-I (Class-Agnostic Query Proposal) provided by the official checkpoint. Then we retrain the model by incorporating uncertainty learning in the more important Stage-II (Class-Specific Segmentation).
The vision encoder uses ResNet-50~\cite{he2016deep} as backbone.
The primary loss $\mathcal{L}_\texttt{occ}$ in VoxFormer consists of the weighted cross entropy loss and scene-class affinity losses on both geometry and semantics from MonoScene~\cite{cao2022monoscene}.
All the models with online uncertainty estimation have been trained with a learning rate of $2e^{-4}$ and a batch size of $1$ per GPU for $20$ epochs.

\textbf{SGN Framework.}
SGN is an end-to-end framework with a ``dense-sparse-dense'' design, which introduces hybrid guidance and effective voxel aggregation. The ResNet-50~\cite{he2016deep} is also adopted as vision backbone. 
The 2D-to-3D transformation is similar to MonoScene~\cite{cao2022monoscene}.
In addition to the loss from VoxFormer, $\mathcal{L}_\texttt{occ}$ in SGN also includes geometric and semantic guidance losses specially designed for the ``dense-sparse-dense'' structure. The related models with online uncertainty estimation are trained with a learning rate of $2e^{-4}$ and a batch size of $1$ per GPU for $40$ epochs.

\subsection{Offline Model Calibration}
In our offline setting, the parameters of the original occupancy network are entirely frozen, with only the parameters in the calibration function and uncertainty learning module being trainable. For the implementation of DeptS in the occupancy network, we calculate the coordinates and depth of the voxel centers to replace the depth of the point cloud used in the original code~\footnote{\url{https://github.com/ldkong1205/Calib3D}}. We adopt a simple log-likelihood loss for $\mathcal{L}_{\texttt{calib}}$.

\section{Additional Experiments}
\label{sec:supp_exp}
\subsection{Qualitative Comparison}
\begin{figure}[t]
\begin{center}
\includegraphics[width=1.0\linewidth]{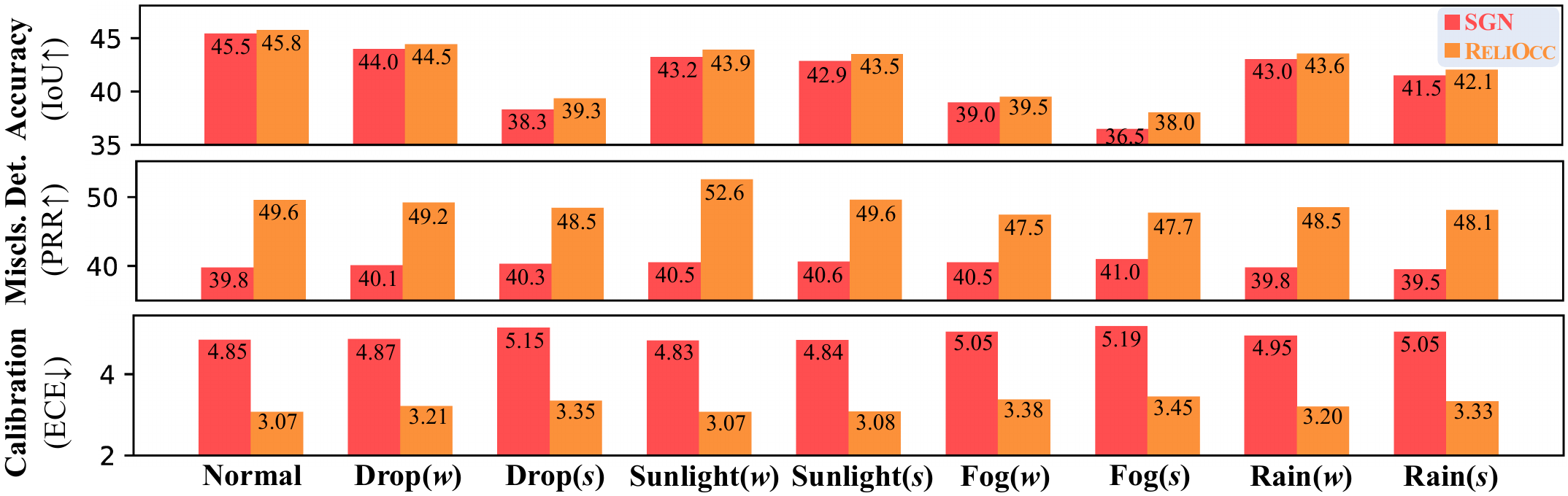} 
\end{center}
\vspace{-2mm}
\caption{The geometric comparison in accuracy and reliability performance between SGN~\cite{mei2023camera} and \model under four out-of-domain conditions.}
\vspace{-3mm}
 \label{fig:robust_supp}
 \end{figure}

We further provide evaluation results on \textit{class-agnostic} geometric metrics from our robustness analysis experiments (see \S\ref{sec:robust_abl}) in Figure \ref{fig:robust_supp}.
Under four different out-of-domain perturbations, our method demonstrates superior geometric reliability ($\texttt{PRR}_{\texttt{geo}}$ and $\texttt{ECE}_{\texttt{geo}}$) compared to SGN~\cite{mei2023camera} and shows significant advantages in accuracy (IoU).

 \begin{figure}[t]
\begin{center}
\includegraphics[width=1.0\linewidth]{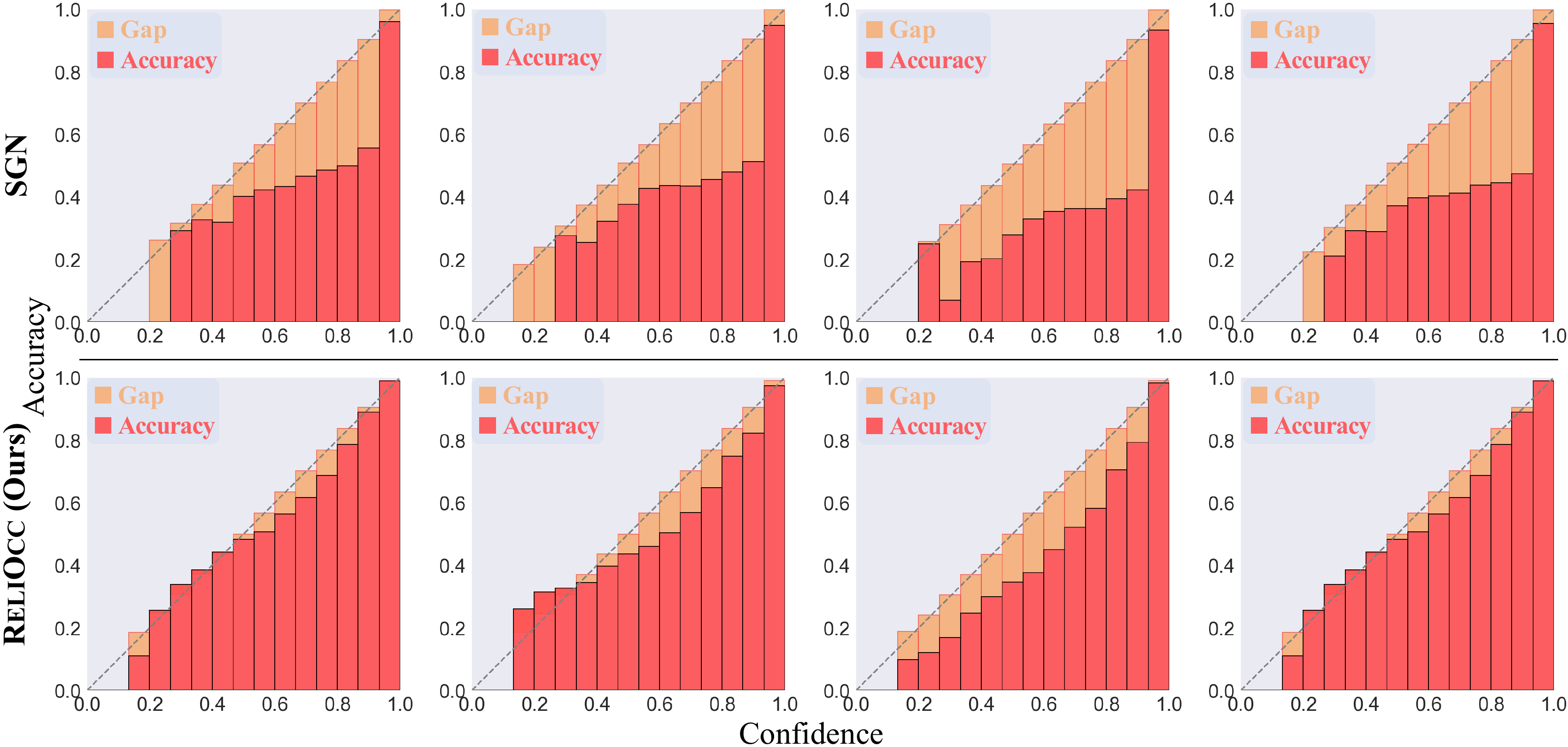} 
\end{center}
\vspace{-2mm}
\caption{More reliability diagrams of semantic calibration gaps in randomly selected predictions from SGN~\cite{mei2023camera} and \model.}
\vspace{-4mm}
 \label{fig:reli_supp}
 \end{figure}

 \begin{figure}[t]
\begin{center}
\includegraphics[width=1.0\linewidth]{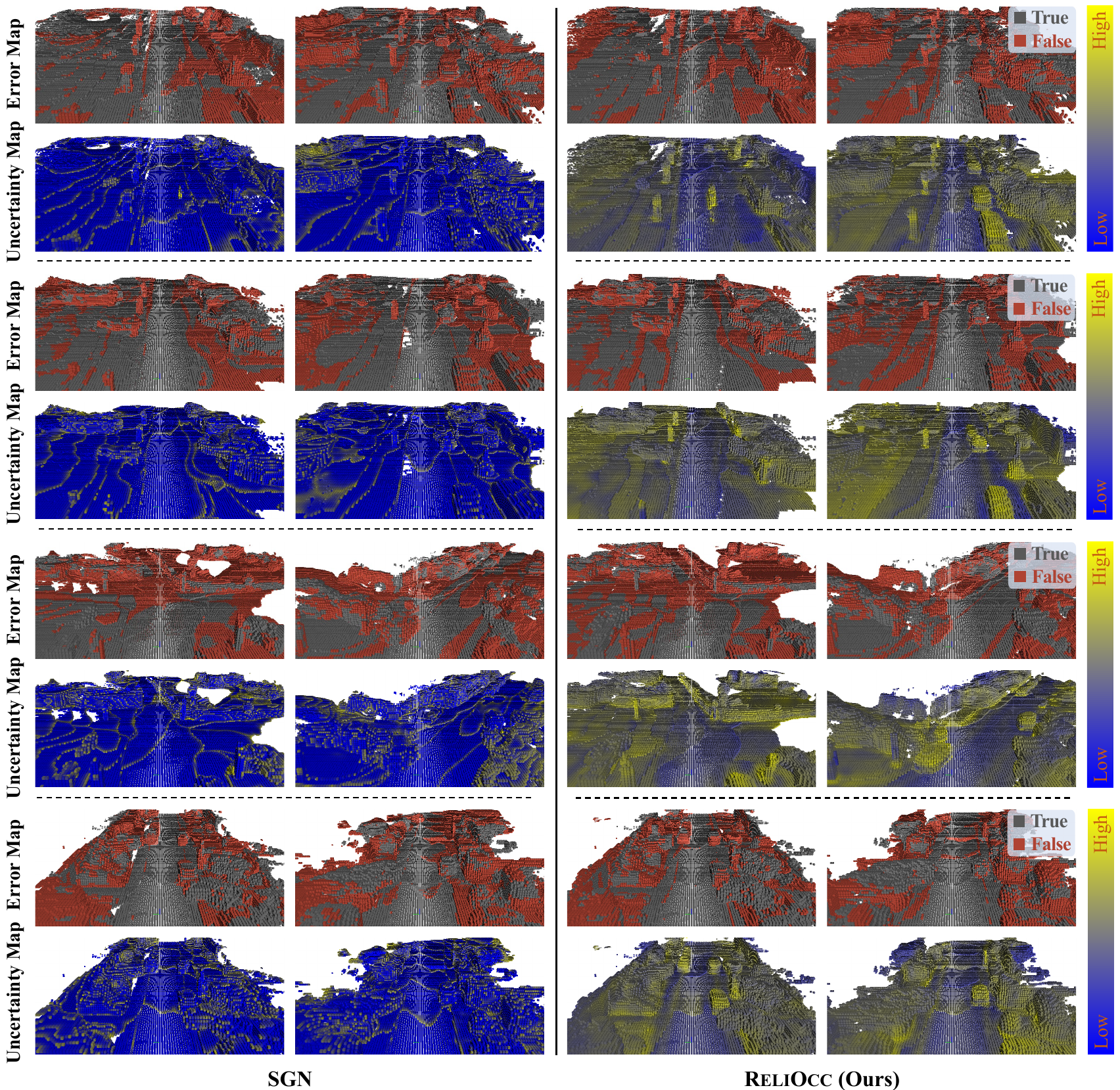} 
\end{center}
\vspace{-3mm}
\caption{More visual results of the error map and corresponding uncertainty map from the predictions by SGN~\cite{mei2023camera} and \model.}
 \label{fig:supp_unc}
 \end{figure}
 
\subsection{Qualitative Comparison}
\textbf{Online Uncertainty Learning.}
We present more error maps with uncertainty maps from the predictions of SGN~\cite{mei2023camera} and \model under the online setting.
As shown in the left column of Figure \ref{fig:supp_unc}, SGN exhibits low uncertainty in many parts of the scene where it makes incorrect predictions, indicating a severe issue of over-confidence. In contrast, in the right column of Figure \ref{fig:supp_unc}, our method demonstrates high uncertainty in most of the incorrect prediction areas, further illustrating that \model effectively addresses the problem of over-confidence and improves model reliability.

\textbf{Offline Model Calibration}
More reliability diagrams of the predictions on the validation set of SemanticKITTI are presented in Figure \ref{fig:reli_supp}.
In various scenarios, our uncertainty-aware calibration strategy effectively reduces calibration gap compared to the SGN~\cite{mei2023camera}, providing more reliable semantic occupancy prediction results.

\section{Further Discussions}
\label{sec:supp_dis}
\subsection{Limitation and Future Work}
\label{sec:supp_dis_limit}
Although our method has achieved significant improvements in reliability metrics for camera-based occupancy models, even approaching the performance of some LiDAR-based methods~\cite{yan2021sparse, xia2023scpnet} in \texttt{PRR} and \texttt{ECE}, we cannot ignore the accuracy gap (mIoU and IoU) between camera and LiDAR. Therefore, camera-based models still have a long way to go. In the future, we will explore using the learned uncertainty in camera to promote label accuracy in offline annotation, reducing reliance on LiDAR and contributing a data loop for a purely vision-based occupancy network.

\subsection{Potential Societal Impact}
\label{sec:supp_dis_social}
Our work explores and enhances the reliability of vision-based occupancy networks, which can significantly improve the safety and accessibility of autonomous driving systems with low-cost sensors. This advancement not only fosters safer navigation and decision-making in autonomous vehicles but also promotes wider adoption of affordable and reliable AI technologies in various applications, contributing positively to transportation efficiency and public safety.
Conversely, the widespread utilization of sensors in autonomous vehicles raises apprehensions regarding data privacy and security. The formulation of robust data storage and management strategies becomes imperative.

\section{License and Consent with Public Resources}
\label{sec:supp_license}
\subsection{Public Datasets}
We use the semantic scene completion annotations provided by the SemanticKITTI dataset and the original image data from the KITTI Odometry Benchmark:
\begin{itemize}
    \item SemanticKITTI\footnote{\url{http://semantic-kitti.org}.} \dotfill CC BY-NC-SA 4.0
    \item SemanticKITTI-API\footnote{\url{https://github.com/PRBonn/semantic-kitti-api}.} \dotfill MIT License
    \item KITTI Odometry Benchmark\footnote{\url{https://www.cvlibs.net/datasets/kitti/eval_odometry.php}.} \dotfill CC BY-NC-SA 3.0
\end{itemize}

\subsection{Re-evaluated Methods}
In the experimental section, we evaluated 10 existing semantic occupancy prediction models, all reproduced using the official code:
\begin{itemize}
    \item SSCNet\footnote{\url{https://github.com/shurans/sscnet}.} \dotfill MIT License
    \item LMSCNet\footnote{\url{https://github.com/astra-vision/LMSCNet}.} \dotfill Apache License 2.0
    \item JS3C-Net\footnote{\url{https://github.com/yanx27/JS3C-Net}.} \dotfill MIT License
    \item SSC-RS\footnote{\url{https://github.com/Jieqianyu/SSC-RS}.} \dotfill MIT License
    \item SCPNet\footnote{\url{https://github.com/SCPNet/Codes-for-SCPNet}.} \dotfill Unknown
    \item MonoScene\footnote{\url{https://github.com/astra-vision/MonoScene}.} \dotfill Apache License 2.0
    \item TPVFormer\footnote{\url{https://github.com/wzzheng/tpvformer}.} \dotfill Apache License 2.0
    \item VoxFormer\footnote{\url{https://github.com/NVlabs/VoxFormer}.} \dotfill NVIDIA Source Code License-NC
    \item NDCScene\footnote{\url{https://github.com/Jiawei-Yao0812/NDCScene}.} \dotfill Unknown
    \item SGN\footnote{\url{https://github.com/Jieqianyu/SGN}.} \dotfill Unknown
    
\end{itemize}

\subsection{Other Resources}
For the calculation of \texttt{PRR} and \texttt{ECE} metrics, the design of out-of-domain perturbations, and the plotting of reliability diagrams, we have referred to the following publicly available implementations for other tasks and made modifications specific to semantic occupancy prediction:
\begin{itemize}
    \item relis\footnote{\url{https://github.com/naver/relis}.} \dotfill CC BY-NC-SA 4.0
    \item 3D Corruptions AD\footnote{\url{https://github.com/thu-ml/3D_Corruptions_AD}.} \dotfill MIT License
    \item reliability diagrams\footnote{\url{https://github.com/hollance/reliability-diagrams}.} \dotfill MIT License
\end{itemize}

\end{document}